\documentclass[review,5p,twocolumn]{elsarticle}

\usepackage{amssymb}
\usepackage{multirow}
\usepackage{amsmath,amsfonts}
\usepackage{algorithmic}
\usepackage{algorithm}
\usepackage{array}
\usepackage[caption=false,font=normalsize,labelfont=sf,textfont=sf]{subfig}
\usepackage{textcomp}
\usepackage{stfloats}
\newcommand{\surname}[1]{#1}
\usepackage{url}
\usepackage{verbatim}
\usepackage{graphicx}
\usepackage[colorlinks,linkcolor=blue]{hyperref}
\usepackage{epstopdf}
\usepackage{multirow}
\usepackage{booktabs}
\usepackage{silence}
\usepackage{bm}
\usepackage{amsmath, bm}

\journal{elsarticle}

\begin{document}

\begin{frontmatter}


\title{Cross-Layer Feature Self-Attention Module for Multi-Scale Object Detection}

\author[1]{Dingzhou \surname{Xie}}
\ead{dingzhouxie@163.com}

\author[3]{Rushi \surname{Lan}}
\ead{rslan2016@163.com}

\author[3]{Cheng \surname{Pang}\corref{cor1}}
\ead{pangcheng3@guet.edu.cn}

\author[2]{Enhao \surname{Ning}}
\ead{ningenhao@xs.ustb.edu.cn}

\author[2]{Jiahao \surname{Zeng}}
\ead{zeng_jiah@163.com}

\author[4]{Wei \surname{Zheng}\corref{cor1}}
\ead{zhengwei@jit.edu.cn}


\cortext[cor1]{Corresponding author: 
}

\affiliation[1]{organization={Guangzhou Huashang College},
                addressline={Hua shang Road}, 
                city={Guangzhou}, 
                postcode={511300}, 
                state={Guangdong},
                country={China}} 

\affiliation[2]{organization={Guangxi Normal University},
                addressline={Yu cai Road}, 
                city={Guilin}, 
                postcode={541004}, 
                state={Gangxi},
                country={China}}

\affiliation[3]{organization={Guilin University of Electronic Technology},
                addressline={Jin ji Road}, 
                city={Guilin}, 
                postcode={541004}, 
                state={Gangxi},
                country={China}}

\affiliation[4]{organization={Jinling Institute of Technology},
                addressline={Hong jing Road}, 
                city={Nanjing}, 
                postcode={211169}, 
                state={Jiangsu},
                country={China}}

\begin{abstract}
Recent object detection methods have made remarkable progress by leveraging attention mechanisms to improve feature discriminability. However, most existing approaches are confined to refining single-layer or fusing dual-layer features, overlooking the rich inter-layer dependencies across multi-scale representations. This limits their ability to capture comprehensive contextual information essential for detecting objects with large scale variations. In this paper, we propose a novel Cross-Layer Feature Self-Attention Module (CFSAM), which holistically models both local and global dependencies within multi-scale feature maps. CFSAM consists of three key components: a convolutional local feature extractor, a Transformer-based global modeling unit that efficiently captures cross-layer interactions, and a feature fusion mechanism to restore and enhance the original representations. When integrated into the SSD300 framework, CFSAM significantly boosts detection performance, achieving 78.6\% mAP on PASCAL VOC (vs. 75.5\% baseline) and 52.1\% mAP on COCO (vs. 43.1\% baseline), outperforming existing attention modules. Moreover, the module accelerates convergence during training without introducing substantial computational overhead. Our work highlights the importance of explicit cross-layer attention modeling in advancing multi-scale object detection.
 \end{abstract}


\end{frontmatter}


\section{Introduction}   
The escalating demand for intelligent automation and industrial inspection has positioned robust multi-scale object detection as a cornerstone technology. While deep learning has driven significant progress, detecting objects with extreme scale variation remains a formidable challenge, primarily due to the semantic and resolution gaps between features at different levels of the network hierarchy \cite{liu2023survey,fang2023multi}. Effectively integrating these multi-scale features to achieve consistent performance across small, medium, and large objects is thus a critical and ongoing research focus.

Attention mechanisms have become instrumental in advancing this field by enabling models to dynamically focus on more informative spatial regions or feature channels \cite{MobileViT,fang2025weathercycle, yang2022focal}. A pivotal aspect of this is modeling long-range dependencies to capture global contextual information. Two dominant paradigms for this purpose are structured convolutional operators \cite{xiao2025fbrt, wang2023internimage,shi2025mamba,wang2025fast,wang2024effective} and self-attention mechanisms \cite{han2022survey,  zhu2025exploring}. While enhanced convolutions can enlarge the receptive field, their capacity for global modeling is often constrained by local connectivity and may require careful hyperparameter tuning to avoid artifacts. In contrast, self-attention mechanisms, catalyzed by the Vision Transformer (ViT) \cite{liu2022convnet}, inherently model pairwise interactions across all positions in a feature map, providing superior flexibility in building global context. Subsequent architectures like the Swin Transformer \cite{Swin-Transformer} have further enhanced their efficiency and applicability to vision tasks.

Notwithstanding their achievements, these methods have some limits in multi-scale object detection. Convolution-based methods often struggle to capture dependencies outside their fixed receptive field. Also, many modern self-attention modules are made for single-scale feature maps \cite{dosovitskiy2020image,hatamizadeh2023global,fang2025guided,hu2023transformer}. They process patches independently and lack a clear way to model connections across different layers of a feature pyramid. This restricts the integration of contextual information across semantic levels and hinders the model from jointly using fine details and high-level semantics, which is very important for accurate multi-scale detection.

To bridge this gap, we propose a Cross-Layer Feature Self-Attention Module (CFSAM). CFSAM is designed to holistically model interactions across multi-scale features within a unified and efficient self-attention framework. It comprises three components: 1) a local feature extractor using convolutional layers to preserve spatial details, 2) a global feature modeling unit that employs a novel partition-based Transformer to efficiently capture long-range dependencies across all input scales simultaneously, and 3) a feature fusion and restoration unit that coherently integrates the refined features back to their original dimensions. The principal contributions of this work are:

\begin{itemize}

\item To effectively exploit long-range dependencies between multi-layer feature maps, we designed the SSD300-CFSAM model. This design enables the model to better understand both the relationships between objects in an image and the contextual information of the scene.
\item The CFSAM module achieves notable performance improvements without a significant increase in model complexity or the number of parameters. Furthermore, its plug-and-play design allows for seamless and flexible integration into various existing neural network architectures.
\item Quantitative and qualitative experimental results on PASCAL VOC and COCO datasets show that our method not only achieves significant performance improvements over the baseline but also maintains superior computational efficiency compared to other enhancement modules.

\end{itemize}

The remainder of this paper is structured as follows: Section 2 reviews related work. Section 3 details the CFSAM methodology. Section 4 presents experimental settings, results, and ablation studies. Finally, Section 5 concludes the paper.

\section{ Related work }   

multi-scale object detection requires different attention to targets at different scales in the same image, while the attention mechanism focuses on weighting important information on the feature map, which allows neural networks to pay more attention to or prioritize specific input features, thus improving the performance of visual tasks such as object detection. There are complementarities between the two, so we conduct an in-depth study on how attention mechanisms can facilitate multi-scale object detection. Among them, channel attention and spatial attention are two commonly used attention mechanisms when processing image data.

Channel attention models include SENet \cite{se}, GSoP-Net \cite{gsop}, and SRM \cite{srm}. In 2017, Hu et al. introduced the concept of channel attention and proposed the SENet model. The core idea of this model is to dynamically adjust the importance of each feature channel using Squeeze and Excitation operations. In 2019, Gao et al. proposed the GSoP-Net model, which utilizes the Global Second-Order Pooling (GSoP) Block to model higher-order statistical information. The GSoP Block also consists of Squeeze and Excitation modules. In the Squeeze module, the GSoP Block first reduces the number of channels with a $1\times1$ convolution, then computes the covariance matrix to capture the correlations between different channels. The covariance matrix is then row-normalized to establish connections between channels. By employing second-order pooling, the GSoP Block enhances the ability to gather global information. Also in 2019, Lee et al. introduced the Style-Based Recalibration Module (SRM). The SRM consists of two main parts: style pooling and style integration. The style pooling performs global average pooling and calculates the standard deviation on the input features to extract style features from each channel. The style integration replaces the original fully connected layer with a lightweight channel-wise fully connected layer to reduce computational requirements. It estimates style weights for each channel based on the results of the style pooling and then recalibrates the feature maps using these style weights to amplify or suppress their information.

Classic spatial attention methods have evolved significantly, with recent research focusing on efficiently integrating and enhancing spatial feature representations. For instance, ACmix\cite{ pan2022integration} elegantly combines the local feature extraction power of convolutional operations with the global receptive field of self-attention, offering a unified computational structure that adaptively fuses the outputs of both paradigms. Meanwhile, for dense prediction tasks such as object detection, Deformable Attention (DAttn) \cite{xia2022vision} has emerged as a prominent solution. It avoids the excessive computational cost of global self-attention by allowing a sparse set of key sampling points to dynamically deform based on the input features, thereby enabling the model to focus on more relevant spatial locations efficiently. These methods exemplify the modern trend of building spatially-aware models that are both powerful and computationally feasible.

Convolutional neural networks excel at handling image data due to their properties such as translation invariance, making them widely used in image processing. However, CNNs lack scale and distortion invariance. In 2015, Jaderberg et al. proposed Spatial Transformer Networks, which learn an affine transformation for each input image to improve the model's ability to handle targets in different positions, orientations, and scales. Integrating spatial transformers into neural network models allows the models to automatically focus on the regions of interest, thereby enhancing the accuracy and robustness of the models in object recognition tasks. Subsequent research efforts have achieved even greater success in this area \cite{wang2023internimage, pan2022fast, gu2023mamba,hu2024toward}.

\begin{figure*}[h]
\begin{minipage}[b]{1.0\linewidth}
  \centering
  \centerline{\includegraphics[width=\textwidth]{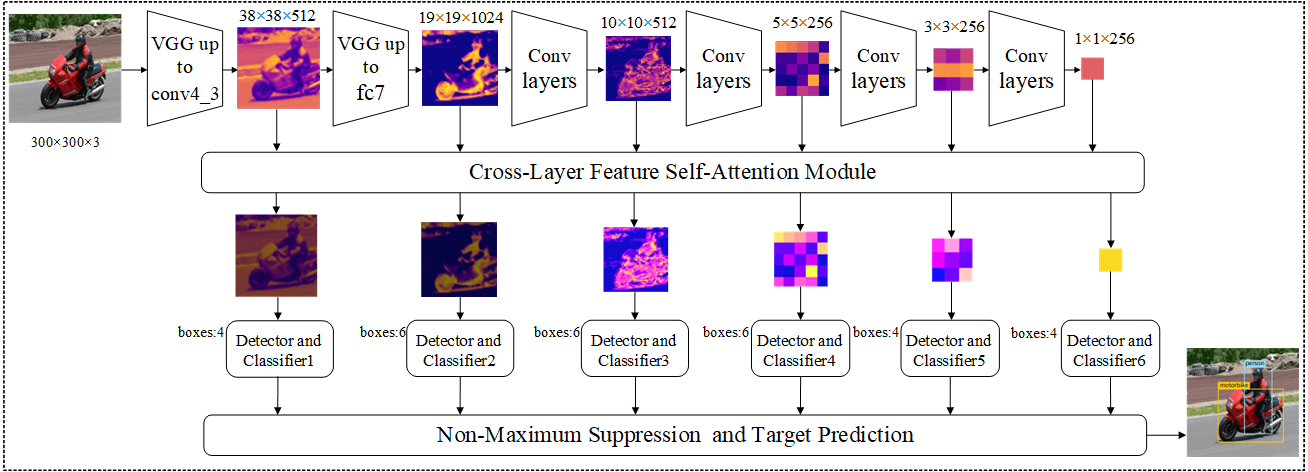}}
\end{minipage}
\caption{Illustration of the whole network. By inserting the CFSAM into the SSD network model, the module takes six predicted feature maps as input. After modeling the dependency relationships, the module outputs the same size and number of predicted feature maps, which are used for object prediction.} 
\label{}

\end{figure*}

The self-attention mechanism is a method for computing attention within a single input sequence. Unlike traditional attention mechanisms \cite{bam,hu2025exploiting,zeng2025explicit,fang2025color}, self-attention considers not only the relevance between each position in the input sequence and a target but also the relevance between each position and other positions. This allows self-attention to capture interactions between all parts of the input sequence, better handling long-range dependencies. In 2017, Vaswani et al. \cite{transformer} proposed the Transformer model, which adopts self-attention to model relationships between different positional elements in the input sequence. The Transformer achieved impressive results in natural language processing tasks, enabling long sequence modeling and parallel computation. Besides natural language processing, the Transformer model has also been applied in computer vision, audio processing, medical diagnosis, and other fields \cite{fcb,zhao2025learning,xing2024personalized,shi2025swimvg}. In 2020, Dosovitskiy et al. introduced the Vision Transformer model, which consists of an encoder composed of multiple Transformer modules and a linear layer for classification. Unlike traditional CNNs, ViT takes a two-dimensional image as input but transforms it into a one-dimensional sequence to perform self-attention computations on the sequence, learning relationships between different positions. To prevent information loss, ViT adds trainable position embeddings at each position of the sequence to provide positional information. The emergence of ViT offers a new perspective in the field of deep learning, applying self-attention to other domains.

Using the core ideas of these methods, we propose to construct a multi-scale object detection model based on cross-layer feature self-attention module to strengthen the ability to model the relationship between feature maps at different levels and enhance the accuracy of multi-scale object detection.

\section{ Methodology }   
Multi-scale object detection networks are deep neural network models used for object detection. They utilize multi-scale prediction feature maps, which have the advantage of effectively detecting objects of different sizes. Common multi-scale object detection networks include Faster R-CNN \cite{fasterrcnn}, YOLO \cite{yolov1, yolov2, yolov3}, and SSD \cite{ssd}. The SSD model predicts objects using anchor boxes of different sizes and aspect ratios on multi-scale feature maps. This reduces the sensitivity of the SSD network to changes in object sizes and improves detection accuracy. We have chosen the SSD object detection network as the baseline model with an input image size of $300\times300$. Figure 1 illustrates how the CFSAM is applied in the SSD network. For multi-scale object detection networks, this module provides a plug-and-play solution.

\begin{figure*}[h]
\begin{minipage}[b]{1.0\linewidth}
  \centering
  \centerline{\includegraphics[width=\textwidth]{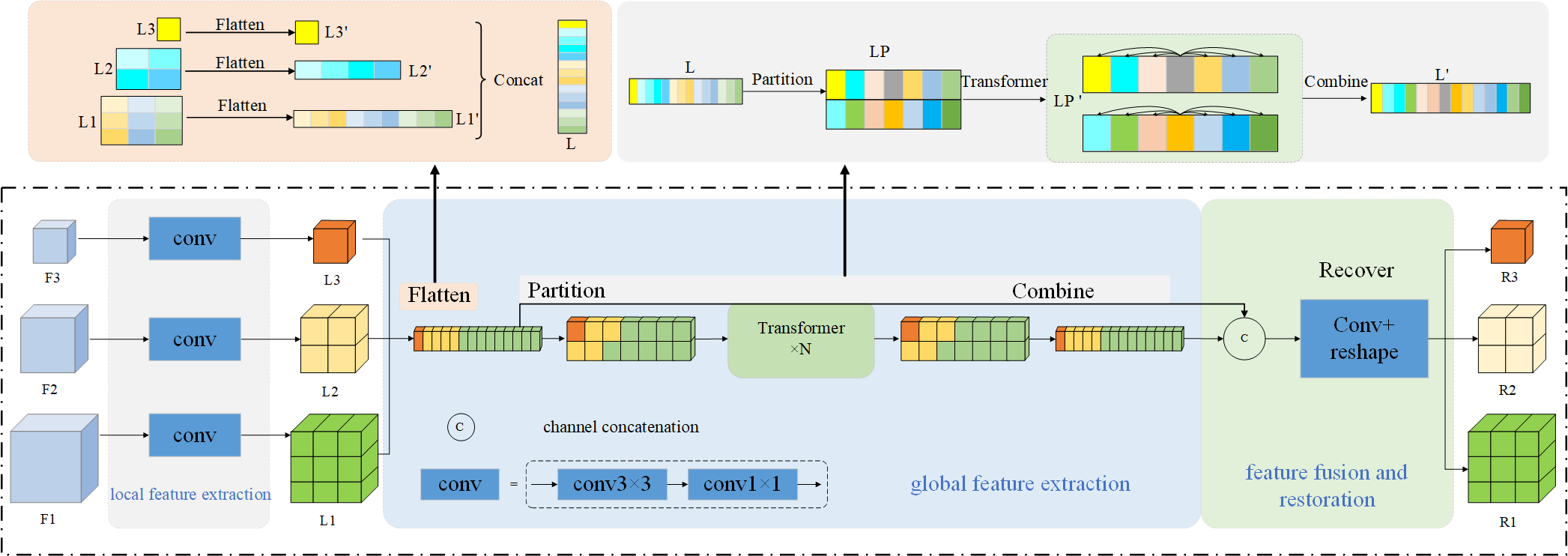}}
\end{minipage}
\caption{Illustrates the structural diagram of the CFSAM, which includes three parts: local feature extraction, global feature extraction, and feature fusion restoration.} 
\label{}

\end{figure*}

The CFSAM we propose consists of three parts: local feature extraction, global feature extraction, and feature fusion restoration. As shown in Figure 2, F1, F2, and F3 represent the multi-scale predicted feature maps. The local feature extraction part utilizes convolutional operations. Specifically, it first performs local feature extraction using a convolutional layer with a kernel size of $3\times3$, and then adjusts the channel count using a convolutional layer with a kernel size of $1\times1$, resulting in feature maps L1, L2, and L3. The global feature extraction part employs Flatten, Partition, and Combine operations for global feature modeling. The Flatten operation is used to flatten the $H\times W$ regions of L1, L2, and L3 feature maps into vectors and concatenate them along the vector dimension to obtain the cross-layer fused feature vector. The Partition operation divides the feature vector into two parts by sampling with a certain interval. During the self-attention calculation, each feature point of a feature vector only interacts with other feature points within itself, reducing the computational complexity. The Combine operation is applied to restore the feature vector back to its original shape. In the feature fusion restoration part, a shortcut branch is first used to concatenate the feature vector obtained after global feature extraction with the cross-layer feature vector before global feature extraction along the channel dimension, doubling the number of channels. Then, a convolutional layer with a kernel size of $1\times1$ is used to adjust the channel count back to the original size. Finally, the feature vector is split and restored to three feature maps R1, R2, and R3, which have the same scale as F1, F2, and F3, respectively. The Transformer unit can capture long-distance dependency relationships in the input sequence and weight the information at each position. Therefore, the resulting three feature maps contain richer contextual information.

\subsection{\textbf{Local feature extraction}} 
Local features are crucial for object classification. For example, when detecting an image of a cat, convolutional layers may recognize local features such as the ears, eyes, and nose of the cat. These local features have a significant impact on determining the target category in the image. To capture the local information of the input feature map, a local feature extraction structure is designed using convolutional layers with kernel sizes of $1\times1$ and $3\times3$. This design effectively promotes the detection efficiency of the target detection model while adding only a small amount of computation.

For the given input feature maps F1, F2, and F3, their channel numbers and sizes are different. They are separately input into a convolutional layer with a kernel size of $3\times3$ and a convolutional layer with a kernel size of $1\times1$. The $3\times3$ convolutional layer is used to extract local spatial features from the image, including edge and texture information. These local features are crucial for object classification. For example, when detecting an image of a cat, the convolutional layer may identify local features such as ears, eyes, and nose, which greatly influence the determination of the object category in the image. The $1\times1$ convolutional layer reduces the channel number of the input feature maps, thereby reducing computational costs and improving detection efficiency. Since the predicted feature maps have a significant difference in channel numbers, with some reaching 1024 dimensions while others have only 256 dimensions, inputting them directly into the Transformer unit without processing would result in higher computational and memory costs for feature maps with larger channel numbers. This would slow down the network's operation speed. Therefore, using a $1\times1$ convolutional layer to uniformly adjust the channel number to 256 dimensions significantly reduces the computational load. Experimental results have shown that the detection accuracy of the network is not compromised. After extracting the local information, output feature maps L1, L2, and L3 are obtained, with the same channel number for all three feature maps.

\subsection{ \textbf{Global feature extraction} }   
The SSD object detection model extracts features of different scales from various layers of the network for prediction, and there is a close relationship between these feature maps at different levels. In object detection tasks, there exists interdependence among different objects. Therefore, in the object detection network, it is necessary to consider the dependency between feature maps at different levels in order to better locate and detect target objects. Before extracting global features, feature fusion is performed. Feature fusion methods include direct addition, channel concatenation, and layer-wise fusion, among others. These methods require upsampling to align the scale of the feature maps, which increases the computational load of the model. We achieve cross-layer fusion of all predicted feature maps by flattening and concatenating them, without using upsampling operations, thereby avoiding an increase in computational load for the model.



As shown in the upper left half of Figure 2, the input feature maps are $L1 \in \mathbb{R}^{H1 \times W1 \times C}$, $L2 \in \mathbb{R}^{H2 \times W2 \times C}$, and $L3 \in \mathbb{R}^{H3 \times W3 \times C}$. Firstly, the feature maps are flattened to $L1' \in \mathbb{R}^{C \times H1W1}$, $L2' \in \mathbb{R}^{C \times H2W2}$, and $L3' \in \mathbb{R}^{C \times H3W3}$. Then, they are concatenated into a feature matrix $L \in \mathbb{R}^{C \times (H1W1+H2W2+H3W3)}$, which preserves a significant amount of low-level feature map information, benefiting the network in detecting small objects. The formula is as follows: 

$$
L=\sum_{i=1}^nFlatten(Conv_{1\times1}(Conv_{3\times3}(F_i))) \eqno{(1)}
$$

By simply flattening and concatenating, we can obtain a cross-layer feature matrix that can effectively save computational resources.



As shown in the upper right half of Figure 2, three operations are involved in extracting global features: Partition, Transformer, and Combine operations. The model needs to input the feature matrix into the Transformer unit to compute attention. If self-attention is computed for each feature point with every other feature point, it would require a significant amount of computational resources. To address this issue, the feature matrix can be divided into a specified number of parts before being input into the Transformer unit, which significantly reduces the computational cost.

First, the Partition operation is performed by setting the value of the partition count. In this case, Part is set to 2, meaning that the feature matrix is split into two feature matrices. By using interval sampling, the feature matrix $L \in R^{C \times Length}$ can be divided into new feature matrices $LP \in R^{Part \times (Length/Part) \times C}$. This splitting method preserves the spatial order of the pixels. If the length of the feature matrix L is not divisible by Part, interpolation is applied to the specified dimension using bilinear interpolation. Compared to the patch-based nonlinear splitting method used in MobileViT Block, the Partition operation enables linear splitting and offers greater flexibility in the splitting approach.The formula is as follows:
$$
LP=Partition(L)  \eqno{(2)}
$$
Next, the Transformer unit is used to model long-distance dependencies. When performing self-attention computation, each feature point in the $LP$ feature matrix only calculates self-attention with the feature points in the same row. This achieves the goal of reducing computational cost. Assuming the height, width, and channel dimensions of the feature map are $H$, $W$, and $C$ respectively, without using the Partition operation, the computational cost is denoted as $O(WHC)$ . After applying the Partition operation with a value of 2 and using interval sampling to select every other feature point within each block for self-attention computation, the computational cost becomes $O(WHC/2)$. Theoretically, the computational cost is reduced by half. This approach is possible because image data itself contains a large amount of data redundancy. For lower-level feature maps, adjacent pixels generally have similar information, and calculating self-attention for each pixel and its neighboring pixels would be computationally wasteful. On larger feature maps, the computational cost of calculating self-attention for neighboring pixels far exceeds the benefits in accuracy. After the Transformer unit computation, we obtain $LP' \in R^{Part \times (Length/Part) \times C}$, which is represented by the following formula:
$$
LP'(p) = Transformer(LP(p)),1 \leqslant p \leqslant Part \eqno{(3)}
$$
Finally, through the Combine operation, the feature matrix $LP'$ is restored to its original arrangement. After this process, we obtain $L' \in R^{C \times Length}$, where $L'$ has the same dimensions as $L$.The formula is as follows:
$$
L^{\prime}=Combine(LP^{\prime}(P)) \eqno{(4)}
$$


\subsection{ \textbf{Feature fusion restoration} }   




In order to obtain richer contextual information and achieve the plug-and-play capability of modules, the reinforced feature vectors are fused with the original feature vectors. By performing convolutional operations, the feature vectors are restored to their original scale without modifying the structure of the target detection network. The processed feature maps have stronger expressive power as they not only capture the local and global relationships between feature maps at different levels but also model the dependencies between different levels and different-sized objects. This improvement in capturing capabilities enhances the performance of the model.


$L \in R^{C \times Length}$ represents the input feature matrix to the Transformer unit, while $L' \in R^{C \times Length}$ represents the output feature matrix from the Transformer unit. Both $L$ and $L'$ have the same scale. By concatenating these two feature maps along the channel dimension, a new feature matrix is generated, providing more feature information. To restore the feature matrix back to the original multi-scale feature maps, a 1x1 convolution layer is used to reduce the channel dimension of the feature matrix, resulting in the feature matrix $R \in R^{C \times Length}$. If bilinear interpolation was used during the splitting of the feature matrix, it is also necessary to use bilinear interpolation here to restore the scale. Then, the feature matrix is sliced and reshaped to obtain three feature maps $R1 \in R^{H1 \times W1 \times C1}$, $R2 \in R^{H2 \times W2 \times C2}$, and $R3 \in R^{H3 \times W3 \times C3}$ with the same scale as the initial feature maps. The formula is as follows:

$$
R_i=Recover(Conv_{1\times1}(Concat(L,L^{\prime}))) \eqno{(5)}
$$
The processed feature maps possess stronger expressive power, capturing the local and global connections between different hierarchical feature maps and modeling the dependencies between different levels and different-sized objects. This helps improve the model's performance.

\section{ Experiments }   

\subsection{ \textbf{Experimental settings} }   
For our experiments, we trained our model on the PASCAL VOC dataset \cite{pascalvoc}, which contains 20 different object classes. The training set consisted of 16,551 images. Our model was implemented using the PyTorch framework and trained and tested on an NVIDIA GeForce RTX 3090 GPU. During the training phase, we used the SGD optimizer to update the parameters of the network model. The batch size was set to 48, and we employed a technique for dynamically adjusting the learning rate. The initial learning rate was set to 1e-3 and was changed to 1e-4 at 130 epochs and 1e-5 at 220 epochs. The momentum parameter was set to 0.9, and the weight decay was set to 5e-4. The total number of training epochs was set to 260.

Also, to better validate our method, we train on the COCO dataset\cite{lin2014microsoft}, which consists of 80 different object classes with more complex object classification scenarios. The dataset contains 82,783 training images and 40,504 test images. Training and testing were performed on NVIDIA GeForce RTX 4060 GPU. The batch size was set to 16, the number of training epochs was set to 120, and the rest of the parameters were set as in the PASCAL VOC dataset.

\begin{figure*}[t]
\begin{minipage}[t]{1.0\linewidth}
  \centering
  \centerline{\includegraphics[width=0.8\textwidth]{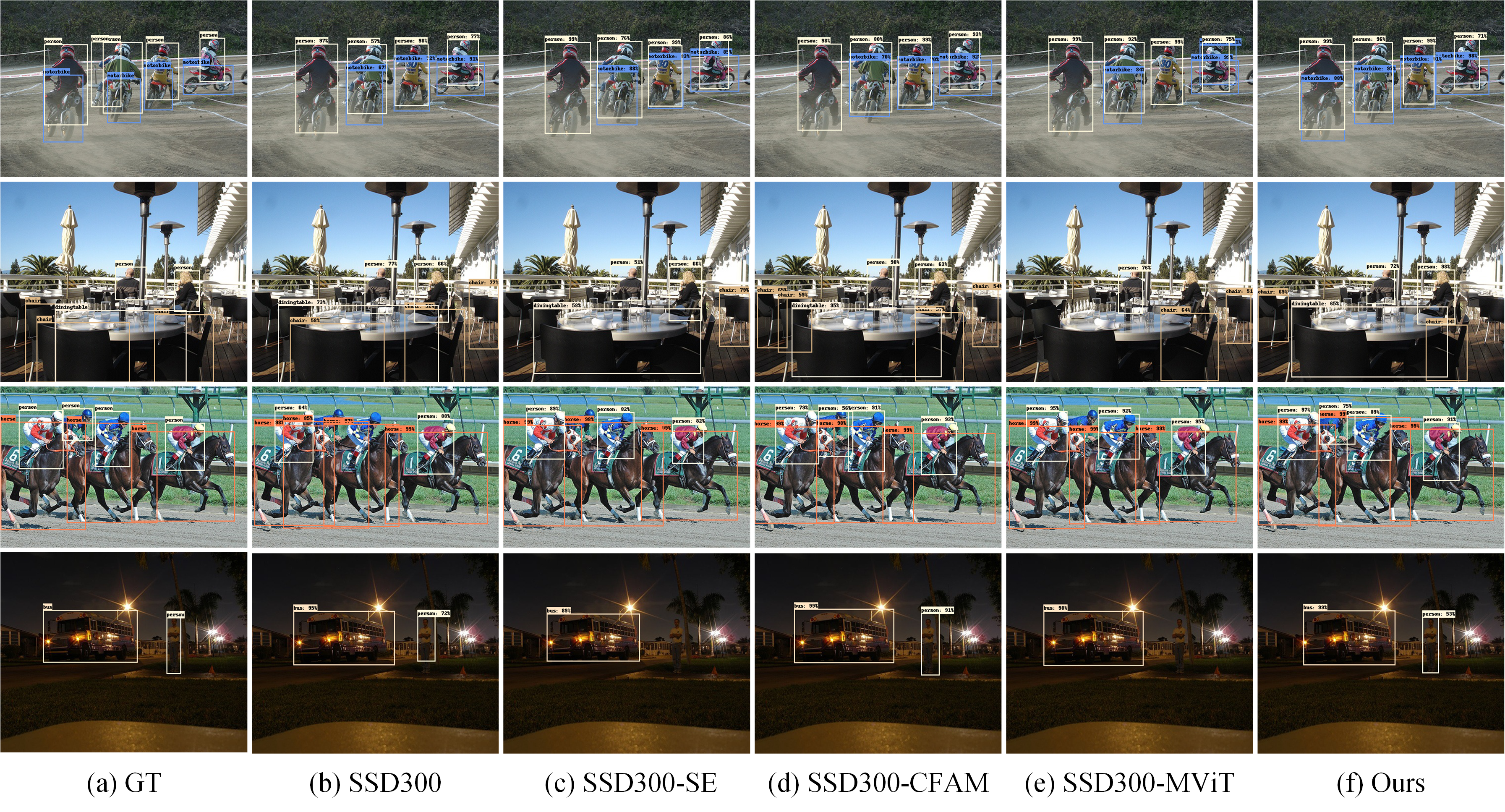}}
\end{minipage}
\caption{Qualitative comparison of the prediction results between our method and other methods on the PASCAL VOC dataset.}
\label{}

\end{figure*}

We selected the SSD multi-scale object detection network as the baseline model. The input image size was set to $300\times300$. We qualitatively and quantitatively compared our proposed modules with SE Block, CFAM \cite{cfam}, and MobileViT Block algorithms. By comparing the differences between various attention mechanisms, we can demonstrate the effectiveness of our proposed method. SE Block controls the contribution of each channel to the final output by computing channel-wise weights, thereby helping the model focus on important channel information and improve accuracy in classification or regression tasks. CFAM explicitly models the interdependencies between different-level feature maps, allowing the network to focus on higher-level feature maps for detecting large objects and lower-level feature maps for detecting small objects. MobileViT Block is a lightweight visual model based on Transformer, designed specifically for resource-constrained scenarios such as mobile devices.

\subsection{ \textbf{Qualitative comparisons} }   

First, we present the prediction result images of different object detection algorithms, comparing them using test images from various scenarios, including dense object images, complex scene object images, and small object images. The results demonstrate that our proposed model is capable of detecting targets with greater accuracy compared to other algorithms (shown in Figure 4).

To visualize the decision process of the neural network, we used the EigenCAM \cite{eigencam} method, which effectively visualizes the contribution of features at different levels of the object detection network to the final detection results. As shown in Figure 5, the results generated by SSD300-SE, SSD300-CFAM, and SSD300-MViT are depicted in Fig. 7(c), (d), and (e) respectively, highlighting the importance of different regions of pixels for the network's detection results. We compared the results using test images from various scenarios, including images with complex scenes, images containing salient objects, and images with dense objects. The first image showcases the detection results in a complex scene, while the second and third images represent the detection of salient objects. The fourth image is an example of detecting dense objects. From the results, it is evident that our proposed model exhibits the most accurate class activation maps compared to other algorithms.

By observing the SSD class activation diagram as in Fig. 7(b), we can find that although the SSD network extracts feature maps at different scales for detection, and the large-scale feature maps can be used to detect small objects while the small-scale feature maps are used to detect large objects; it does not effectively establish the local-global connection of the feature maps at different scales. By introducing the CFSAM module, we can find that the SSD-CFSAM class activation map as in Fig. 7(f) creates a connection between local and global features, and it can be inferred that the CFSAM module effectively enhances the ability of SSD missing.

\begin{table*}[h]
    \centering
    \caption{Quantitative comparison of detection accuracy between our method and other methods on the PASCAL VOC dataset. Bold values indicate the highest AP for each category. }
    \begin{tabular}{@{}cccccccc@{}}
    \hline
Model	& SSD300 & HyperNet	& Ion & SSD300-SE & SSD300-CFAM & SSD300-MViT &	Ours\\
\midrule
mAP	    & 75.5	& 76.3 & 76.5 & 77.2 & 77.6 & 78.1 & 78.6 \\
aero & 81.2	& 77.4 & 79.2 & 83.4 & 79.6 & 84.1 & 82.7 \\
bic. & 82.4	& 83.3 & 79.2 & 83.5 & 84.6 & 85.5 & 85.4 \\
bird & 72.4	& 75.0 & 77.4 & 75.3 & 76.3 & 77.0 & 76.1 \\
boat & 66.0	& 69.1 & 69.8 & 69.9 & 71.8 & 70.0 & 73.3 \\
bot. & 47.2	& 62.4 & 55.7 & 50.0 & 50.6 & 51.6 & 52.7 \\
bus	    & 84.2	& 83.1 & 85.2 & 84.6 & 85.1 & 86.5 & 85.4 \\
car	    & 85.1	& 87.4 & 84.2 & 86.0 & 86.1 & 86.6 & 86.9 \\
cat	    & 85.3	& 87.4 & 89.8 & 86.6 & 89.1 & 88.1 & 88.8 \\
cha. & 57.4	& 57.1 & 57.5 & 60.7 & 62.4 & 62.8 & 62.2 \\
cow	    & 80.9	& 79.8 & 78.5 & 82.1 & 84.6 & 85.3 & 87.2 \\
din. & 76.0	& 71.4 & 73.8 & 76.0 & 75.9 & 74.5 & 76.6 \\
dog	    & 82.1	& 85.1 & 87.8 & 85.5 & 86.2 & 85.8 & 86.1 \\
hor. & 85.9	& 85.1 & 85.9 & 85.7 & 87.6 & 86.7 & 88.0 \\
mot. & 82.9	& 80.0 & 81.3 & 83.4 & 83.3 & 84.0 & 85.1 \\
per. & 77.3	& 79.1 & 75.3 & 78.2 & 77.5 & 79.6 & 79.3 \\
pot. & 49.9	& 51.2 & 49.7 & 53.4 & 50.4 & 51.2 & 52.1 \\
she. & 76.7	& 79.1 & 76.9 & 77.1 & 77.9 & 78.0 & 78.9 \\
sofa & 76.7	& 75.7 & 74.6 & 79.4 & 79.3 & 79.6 & 79.1 \\
tra. & 86.5	& 80.9 & 85.2 & 86.4 & 87.8 & 87.8 & 88.1 \\
tv.	    & 73.8	& 76.5 & 82.1 & 77.1 & 76.3 & 77.1 & 78.1 \\ \hline
    \end{tabular}
    \label{}
\end{table*}

\begin{figure*}[t]
\begin{minipage}[t]{1.0\linewidth}
  \centering
  \centerline{\includegraphics[width=0.8\textwidth]{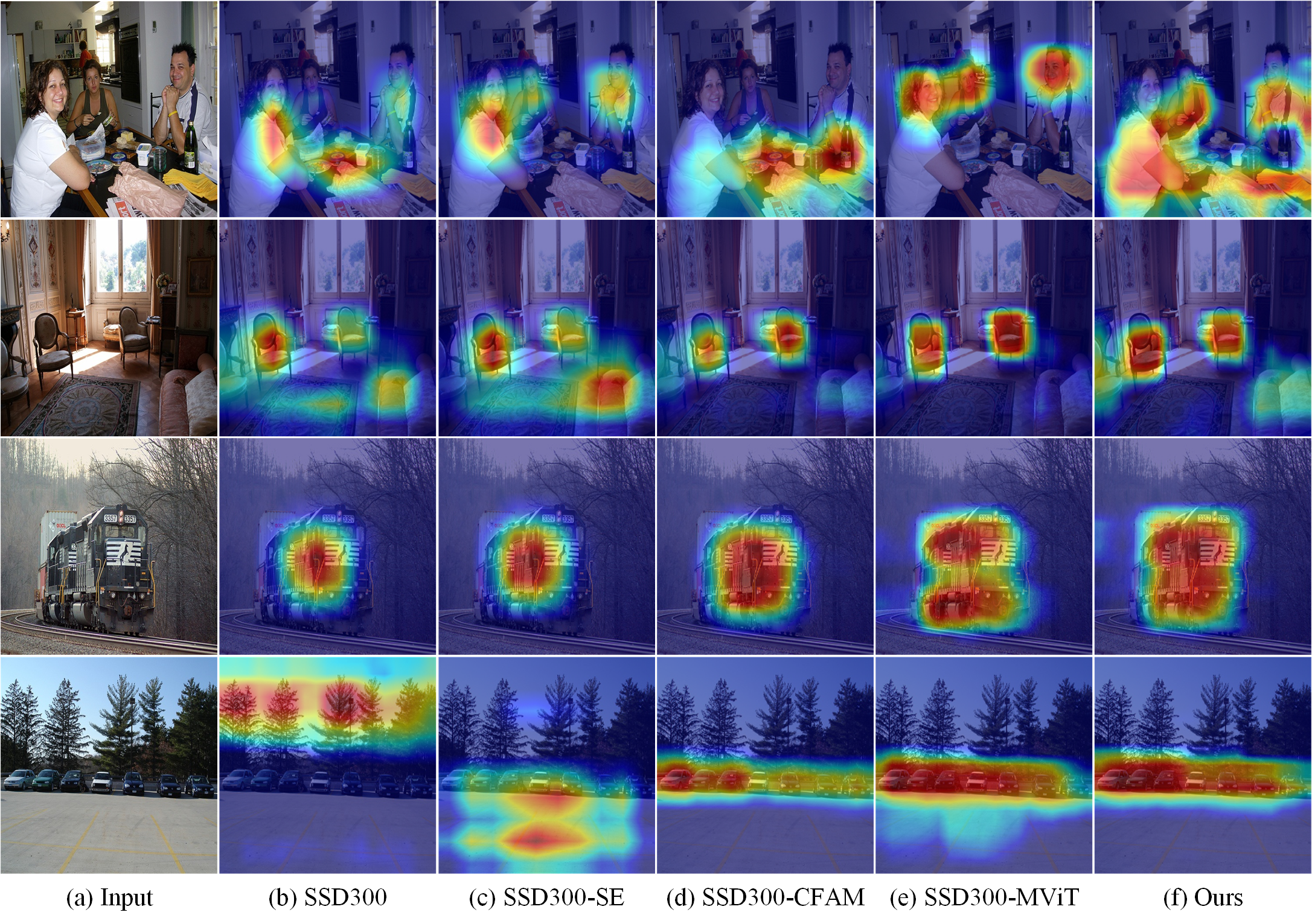}}
\end{minipage}
\caption{Qualitative comparison of class activation maps between our method and other methods on the PASCAL VOC dataset.}
\label{}

\end{figure*}

\subsection{ \textbf{Quantitative comparisons} }   

\begin{table}[h]
    \centering
    \caption{Quantitative comparison of the detection accuracy of our method with SSD-Det and BoundConvNet on PASCAL VOC dataset.}
    \begin{tabular}{@{}ccccc@{}}
    \hline

Model & SSD300 & SSD-Det & BoundConvNet & Ours \\
\midrule
mAP	& 75.5 & 77.1 & 78.5 & 78.6 \\ \hline
    \end{tabular}
    \label{}
\end{table}

\begin{table}[h]
    \centering
    \caption{Quantitative comparison of detection accuracy between our method and other methods on the COCO dataset.}
    \begin{tabular}{@{}ccccc@{}}
    \hline

Model & SSD300 & RCNN-FPN & EfficientDet & Ours \\
\midrule
AP@0.5	& 41.2 & 56.9 & 53.0 & 52.1 \\ \hline
    \end{tabular}
    \label{}
\end{table}

\begin{table}[h]
    \centering
    \caption{Quantitative Comparison of Detection Speed, Computational Complexity, and Parameter Count between Our Method and Other Methods on the PASCAL VOC Dataset.}
    \begin{tabular}{@{}ccccc@{}}
    \hline

Model & FPS↑ & GFLOPs↓ & Params/MB↓ \\
\midrule
SSD300	& 38 &	31.37 &	26.29 \\
SSD300-SE	& 36 &	31.38 &	26.51 \\
SSD300-CFAM	& 44 &	30.93 &	25.69 \\
SSD300-MViT	& 44 & 53.74 &	76.75 \\
Ours & 	57 & 45.22 & 45.16 \\  \hline
    \end{tabular}
    \label{}
\end{table}

We conducted numerical evaluations of different network models on the VOC2007 test set. The evaluation results are shown in Table 1, which presents the quantitative comparison of various methods, including SSD300, HyperNet \cite{hypernet}, Ion \cite{ion}, SSD300-SE, SSD300-CFAM, and SSD300-MViT. Compared to other networks, our proposed model achieves a higher mAP. By introducing the CFSAM module into the SSD300 model, the mAP value is improved by 3.1\%, leading to a significant enhancement in detection performance. As a comparative experiment, the SSD300-MViT model, despite having a significantly higher number of parameters due to the addition of multiple MobileViT Blocks, obtains a lower mAP value. This is because the MobileViT Block focuses on refining individual feature maps without considering the relationships among multiple prediction feature layers, which leads to inferior performance in multi-scale object detection networks compared to the CFSAM module. Table 1 also displays the AP values of these network models for the 20 object categories. From the data in the table, it can be observed that our proposed model outperforms existing methods, particularly in the category of small objects.

Meanwhile, we compare the performance of SSD-Det\cite{wu2023spatial}and BoundConvNet \cite{zhang2023bounding}, the target detection models proposed in recent years, on the VOC2007 dataset, and as shown in Table 2, our proposed SSD-CFSAM outperforms them.

In order to better evaluate our model, we perform numerical evaluation on the COCO2014 test set.The number of image targets in the COCO dataset is higher and the target size is smaller, so the task on the COCO dataset is more challenging. The evaluation results are shown in the table 3, by quantitatively comparing with SSD300, RCNN-FPN\cite{bell2016inside}, and EfficientDet \cite{tan2020efficientdet}.Compared with the baseline, our proposed CFSAM module significantly improves the performance of SSD300 on the COCO dataset, and the AP@0.5 value is improved by 10.9\% compared with the baseline model.Although the SSD model is one of the pioneers in the field of single-stage target detection, with the iteration of target detection technology, it is being surpassed by models such as Faster RCNN-FPN and efficientDet surpassed, as a comparison experiment we can see that our proposed CFSAM module applied to the SSD model closes the distance with Faster RCNN-FPN and EfficientDet-D0 and achieves a good performance.

Table 4 shows the comparison of FPS, GFLOPs, and parameter count for five network models. SSD300-SE and SSD300-CFAM models use channel attention methods, while SSD300-MViT model and our proposed model use self-attention methods, therefore, the SSD300-MViT model and our proposed model based on self-attention mechanism have higher floating-point operations per second (GFLOPs) and more parameters in the Params indicator. Because our model uses the Partition operation to reduce the computation of self-attention, our model has a significant reduction in GFLOPs and Params compared to the SSD300-MViT model, and performs better. Our model has the best performance in the FPS indicator with the fastest detection speed. Although the FPS of object detection models is usually related to the model parameter count and computational complexity, it is not absolute, and many factors affect the FPS of the network model. FPS is more dependent on the speed of forward propagation of the object detection model.

\begin{figure}[t]
\begin{minipage}[t]{1.0\linewidth}
  \centering
  \centerline{\includegraphics[width=0.8\textwidth]{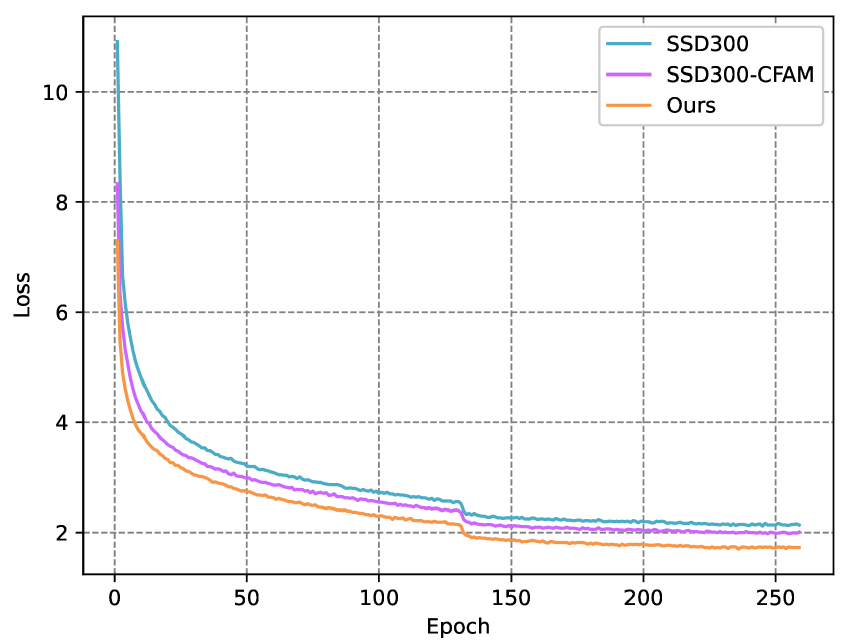}}
\end{minipage}
\caption{Comparison of training loss between our method and other methods on the PASCAL VOC dataset.}
\label{}

\end{figure}

We also compared the training losses of the models. To make a fair comparison, we set the number of training epochs to 260 and used dynamic learning rate adjustment. The initial learning rate was set to 1e-3 and changed to 1e-4 and 1e-5 at 130 and 220 epochs, respectively. As shown in Figure 6, the baseline SSD300 model converged the slowest, while our proposed model converged the fastest. The learning rate adjustment at 130 epochs resulted in a significant decrease in loss value, and our model had the lowest loss value at convergence. These results demonstrate that the CFSAM module can speed up the training process and reduce the convergence loss of the network model.
\subsection{ \textbf{Ablation study} }   
A comprehensive ablation study was conducted to meticulously evaluate the impact of each component within the proposed CFSAM module.  The experiments were performed on the PASCAL VOC dataset using the SSD300 as the baseline model.  To clearly illustrate the contribution of each component, we report the Average Precision (AP) for three representative object categories—bird (small objects), bus (large objects), and cow (medium-sized objects)—in Table 5.  These categories were selected to reflect the model's performance across objects of different scales.

The experiments commenced with the baseline SSD300 model. We then progressively integrated the three core components of CFSAM: Local Feature Extraction (LFE), Global Feature Extraction (GFE), and Feature Fusion Restoration (FFR). The results are encapsulated in Table 5. The baseline model achieved APs of 72.4\%, 84.2\%, and 80.9\% for bird, bus, and cow, respectively. The introduction of the LFE component, responsible for capturing spatial details and unifying channel dimensions, brought noticeable improvements across all categories, with the AP for cow rising to 82.1\%. This underscores the importance of preserving fine-grained local information.

Subsequently, adding the GFE component, which models long-range, cross-layer dependencies via our novel Partition-based Transformer, led to a further performance boost. The AP for the challenging bird category increased to 75.3\%, demonstrating that capturing global contextual relationships is crucial for detecting objects with extreme scale variations, particularly small targets.

The final integration of the FFR component, which coherently fuses the original and enhanced features, resulted in the highest AP scores for all three categories, reaching 76.1\%, 85.4\%, and 87.2\% for bird, bus, and cow, respectively. This confirms that the synergistic operation of all three components enables the model to construct a more powerful and discriminative feature representation, effectively combining local precision with global context.

\begin{table}[h]
    \centering
    \caption{ Ablation study of the proposed CFSAM components on the PASCAL VOC 2007 test set.}
    \begin{tabular}{@{}cccccc@{}}
    \hline
        
Model & bird & bus & cow \\ \hline
        SSD300 (Baseline) &72.4 &84.2 &80.9\\
        SSD300+LEF   &73.2&84.7  &79.4\\
        SSD300+LEF+GFE &75.4  & 84.5  &85.7\\
        SSD300+CFSAM(Ours)  &76.1 &85.4  &87.2\\
 \hline
    \end{tabular}
    \label{}
\end{table}





To further investigate the impact of CFSAM's placement within the feature pyramid, we conduct an ablation study by inserting the module into different subsets of the SSD300 prediction layers.  Specifically, we compare three configurations: applying CFSAM only to the first three layers (shallow features), only to the last three layers (deep features), and to all six prediction layers.  We evaluate the performance on three representative object categories—bird (small objects), bus (large objects), and cow (medium-sized objects)—to analyze the scale-specific effects.  Results on the PASCAL VOC 2007 test set are summarized in Table 6.

\begin{table}[h]
    \centering
    \caption{ Ablation study on the insertion positions of CFSAM in SSD300 on PASCAL VOC 2007 test set. }
    \begin{tabular}{@{}cccccc@{}}
    \hline
        
Model & bird & bus & cow \\ \hline
        SSD300 (Baseline) &72.4 &84.2 &80.9\\
        +CFSAM(Shallow)   &74.8&84.6  &83.2\\
        +CFSAM(Deep) &73.5  & 85.0  &84.7\\
        +CFSAM(All)  &76.1 &85.4  &87.2\\
 \hline
    \end{tabular}
    \label{}
\end{table}

We observe that inserting CFSAM at any level brings improvements over the baseline across all object scales.  However, the full configuration (CFSAM-All) achieves the highest AP scores for all three categories, demonstrating the importance of cross-layer attention across the entire feature hierarchy.  Notably, applying CFSAM only to the first three layers shows the strongest improvement for small objects, while applying it only to the last three layers benefits large and medium-sized objects detection the most.  This aligns with the intuition that shallow features contain more detailed information beneficial for small objects, while deep features capture higher-level semantics crucial for large and medium-sized objects.  Nevertheless, only the complete integration across all six layers maximizes performance uniformly across all object scales, enabling comprehensive modeling of dependencies from fine-grained details to high-level semantics.

\section{ Discussion }     

In this paper we propose a cross-layer feature retention module (CFSAM) and also construct a modified version of the SSD300 multiscale target detection model based on this module. The experimental results verify that the proposed cross-layer feature retention module (CFSAM) with plug-and-play capability is more performant relative to other enhancement modules, and the CFSAM-based multiscale target detection model is greatly improved in the baseline. However, due to the use of a transformer computational unit in CFSAM, the approach cannot be applied in object detection models for mobile devices. In future work, we will explore optimizing CFSAM in terms of reducing feature dimensions and compressing model parameters to address this limitation.






\bibliographystyle{elsarticle-num} 
\bibliography{occ}

\end{document}